%% file: main.tex
\documentclass{article} 
\PassOptionsToPackage{numbers,compress,sort}{natbib}
\usepackage{times}
\usepackage[final]{nips16}
\usepackage{authblk}
\usepackage{hyperref}
\usepackage{url}
\usepackage{graphicx}
\usepackage{amsmath,amssymb} 
\usepackage{pgfplots}
\usepackage{placeins}
\pgfplotsset{compat=newest}

\usepackage{tabularx}

\newcommand{\fig}[1]{Figure~\ref{fig:#1}}

\newcommand{\eq}[1]{(\ref{eq:#1})}

\newcommand{\tS}{\ensuremath{\theta_S}}
\newcommand{\tR}{\ensuremath{\theta_R}}
\renewcommand{\b}{\ensuremath{\mathbf{b}}}
\renewcommand{\r}{\ensuremath{\mathbf{r}}}
\renewcommand{\S}{\ensuremath{\mathcal{S}}}
\newcommand{\R}{\ensuremath{\mathcal{R}}}
\newcommand{\T}{\ensuremath{\mathcal{T}}}
\newcommand{\C}{\ensuremath{\mathcal{C}}}
\newcommand{\sign}{\ensuremath{\mathop{sign}}}

\newcommand{\Exp}{{\mathop{\scalebox{1.2}{\ensuremath{\mathbb{E}}}}\nolimits\,}}

\title{Learnable Visual Markers}

\author[1]{Oleg Grinchuk}
\author[1,2]{Vadim Lebedev}
\author[1]{Victor Lempitsky}
\affil[1]{Skolkovo Institute of Science and Technology\\
Moscow, Russia}
\affil[2]{Yandex\\
Moscow, Russia}

\begin{document}

\maketitle

\begin{abstract}
We propose a new approach to designing visual markers (analogous to QR-codes, markers for augmented reality, and robotic fiducial tags) based on the advances in deep generative networks. In our approach, the markers are obtained as color images synthesized by a deep network from input bit strings, whereas another deep network is trained to recover the bit strings back from the photos of these markers. The two networks are trained simultaneously in a joint backpropagation process that takes characteristic photometric and geometric distortions associated with marker fabrication and marker scanning into account. Additionally, a stylization loss based on statistics of activations in a pretrained classification network can be inserted into the learning in order to shift the marker appearance towards some texture prototype. In the experiments, we demonstrate that the markers obtained using our approach are capable of retaining bit strings that are long enough to be practical. The ability to automatically adapt markers according to the usage scenario and the desired capacity as well as the ability to combine information encoding with artistic stylization are the unique properties of our approach. As a byproduct, our approach provides an insight on the structure of patterns that are most suitable for recognition by ConvNets and on their ability to distinguish composite patterns.
\end{abstract}

\section{Introduction}

Visual markers (also known as visual fiducials or visual codes) are used to facilitate human-environment and robot-environment interaction, and to aid computer vision in resource-constrained and/or accuracy-critical scenarios. Examples of such markers include simple 1D (linear) bar codes~\cite{Woodland52} and their 2D (matrix) counterparts such as QR-codes~\cite{Hara98} or Aztec codes~\cite{Longacre97}, which are used to embed chunks of information into objects and scenes. In robotics, AprilTags~\cite{Olson11} and similar methods~\cite{Scharstein01,Claus04,Bergamasco13} are a popular way to make locations, objects, and agents easily identifiable for robots. Within the realm of augmented reality (AR), ARCodes~\cite{Fiala05} and similar marker systems~\cite{Mooser06,Kaltenbrunner07} are used to enable real-time camera pose estimation with high accuracy, low latency, and on low-end devices. Overall, such markers can embed information into the environment in a more compact and language-independent way as compared to traditional human text signatures, and they can also be recognized and used by autonomous and human-operated devices in a robust way.

Existing visual markers are designed ``manually'' based on the considerations of the ease of processing by computer vision algorithms, the information capacity, and, less frequently, aesthetics.  Once marker family is designed, a computer vision-based approach (a \textit{marker recognizer}) has to be engineered and tuned in order to achieve reliable marker localization and interpretation \cite{Lo95,Richardson13,Belussi13}. The two processes of the visual marker design on one hand and the marker recognizer design on the other hand are thus separated into two subsequent steps, and we argue that such separation makes the corresponding design choices inherently suboptimal. In particular, the third aspect (aesthetics) is usually overlooked, which leads to visually-intrusive markers that in many circumstances might not fit the style of a certain environment and make this environment ``computer-friendly'' at the cost of ``human-friendliness''.

In this work, we propose a new general approach to constructing use visual markers that leverages recent advances in deep generative learning. To this end, we suggest to embed the two tasks of the visual marker design and the marker recognizer design into a single end-to-end learning framework. Within our approach, the learning process produces markers and marker recognizers that are adapted to each other ``by design''. While our idea is more general, we investigate the case where the markers are synthesized by a deep neural network (the \textit{synthesizer} network), and when they are recognized by another deep network (the \textit{recognizer} network). In this case, we demonstrate how these two networks can be both learned by a joint stochastic optimization process. 

The benefits of the new approach are thus several-fold:
\begin{enumerate}
    \item As we demonstrate, the learning process can take into account the adversarial effects that complicate recognition of the markers, such as perspective distortion, confusion with background, low-resolution, motion blur, etc. All such effects can be modeled at training time as piecewise-differentiable transforms. In this way they can be embedded into the learning process that will adapt the synthesizer and the recognizer to be robust with respect to such effect.
    
    \item It is easy to control the trade-offs between the complexity of the recognizer network, the information capacity of the codes, and the robustness of the recognition towards different adversarial effects. In particular, one can set the recognizer to have a certain architecture, fix the variability and the strength of the adversarial effects that need to be handled, and then the synthesizer will adapt so that the most ``legible'' codes for such circumstances can be computed.
    
    \item Last but not least, the aesthetics of the neural codes can be brought into the optimization. Towards this end we show that we can augment the learning objective with a special stylization loss inspired by \cite{Gatys15,Gatys16,Ulyanov16}. Including such loss facilitates the emergence of stylized neural markers that look as instances of a designer-provided stochastic texture. While such modification of the learning process can reduce the information capacity of the markers, it can greatly increase the ``human-friendliness'' of the resulting markers.
\end{enumerate}
Below, we introduce our approach and then briefly discuss the relation of this approach to prior art. We then demonstrate several examples of learned marker families.

\section{Learnable visual markers}
\begin{figure}
    \centering
    \includegraphics[width=\textwidth]{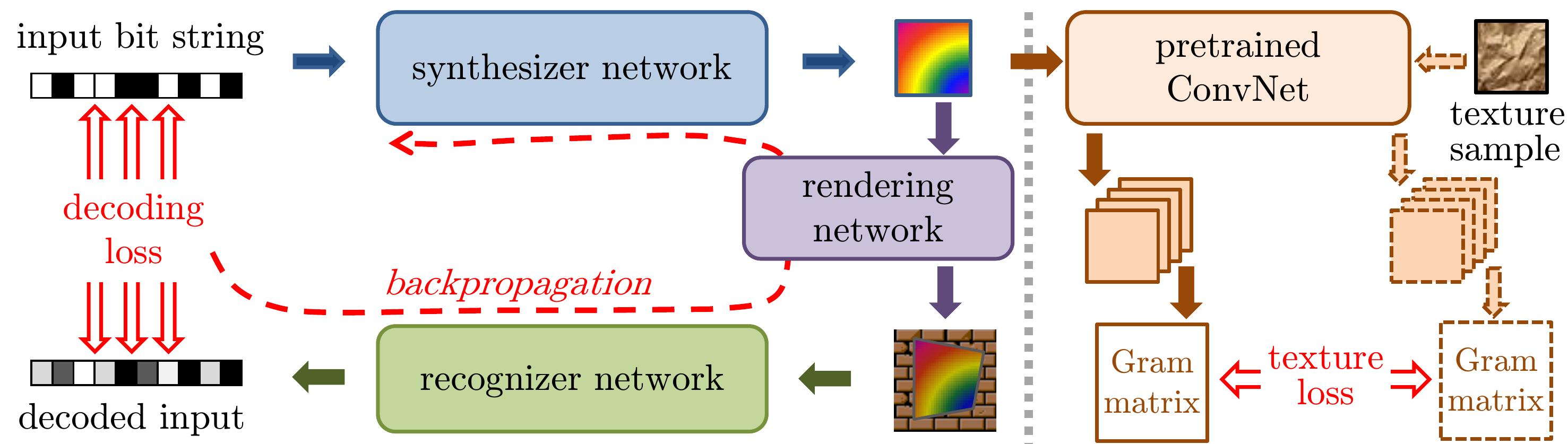}
    \caption{The outline of our approach and the joint learning process. Our core architecture consists of the \textit{synthesizer network} that converts input bit sequences into visual markers, the \textit{rendering network} that simulates photometric and geometric distortions associated with marker printing and capturing, and the \textit{recognizer network} that is designed to recover the input bit sequence from the distorted markers. The whole architecture is trained end-to-end by backpropagation, after which the synthesizer network can be used to generate markers, and the recognizer network to recover the information from the markers placed in the environment. Additionally, we can enforce the visual similarity of markers to a given texture sample using the mismatch in deep Gram matrix statistics in a pretrained network \cite{Gatys15} as the second loss term during learning (right part). }
    \label{fig:scheme}
\end{figure}

We now detail our approach (\fig{scheme}). Our goal is to build a \textit{synthesizer network} $\S(\b;\tS)$ with learnable parameters $\tS$ that can encode a bit sequence $\b=\{b_1,b_2,\dots b_n\}$ containing $n$ bits into an image $M$ of the size $m$-by-$m$ (a \textit{marker}). For notational simplicity in further derivations, we assume that $b_i \in \{-1,+1\}$.

To recognize the markers produced by the synthesizer, a \textit{recognizer network} $\R(I;\tR)$ with learnable parameters $\tR$ is created. The recognizer takes an image $I$ containing a marker and infers the real-valued sequence $\r = \{r_1,r_2,\dots,r_n\}$. The recognizer is paired to the synthesizer to ensure that $\sign r_i = b_i$, i.e.\ that the signs of the numbers inferred by the recognizers correspond to the bits encoded by the synthesizer. In particular, we can measure the success of the recognition using a simple loss function based on element-wise sigmoid:
\begin{equation}\label{eq:loss}
L(\b,\r) = -\frac{1}{n}\sum_{i=1}^n \sigma( b_i r_i ) = -\frac{1}{n}\sum_{i=1}^n \frac{1}{1+\exp(-b_i r_i)}\,
\end{equation}
where the loss is distributed between $-1$ (perfect recognition) and $0$.

In real life, the recognizer network does not get to work with the direct outputs of the synthesizer. Instead, the markers produced by the synthesizer network are somehow embedded into an environment (e.g.\ via printing or using electronic displays), and later their images are captioned by some camera controlled by a human or by a robot. During learning, we model the transformation between a marker produced by the synthesizer and the image of that marker using a special feed-forward network (the\textit{ renderer network}) $\T(M;\phi)$, where the parameters of the renderer network $\phi$ are sampled during learning and correspond to background variability, lighting variability, perspective slant, blur kernel, color shift/white balance of the camera, etc. In some scenarios, the \textit{non-learnable} parameters $\phi$ can be called nuisance parameters, although in others we might be interested in recovering some of them (e.g.\ the perspective transform parameters). During learning $\phi$ is sampled from some distribution $\Phi$ which should model the variability of the above-mentioned effects in the conditions under which the markers are meant to be used.

When our only objective is robust marker recognition, the learning process can be framed as the minimization of the following functional:
\begin{equation} \label{eq:obj1}
f(\tS,\tR) = \Exp_{\substack{\b \sim U(n) \\ \phi \sim \Phi}}\, L\left(\,\b,\, \R\left(\,\T\left(\S(\b;\tS);\,\phi \strut \right)\,;\,\tR\rule{-1pt}{2.2ex}\right)\, \rule{-1pt}{3ex}  \right)\,.
\end{equation}
Here, the bit sequences $\b$ are sampled uniformly from $U(n) = \{-1;+1\}^n$, passed through the synthesizer, the renderer, and the recognizer, with the (minus) loss \eq{loss} being used to measure the success of the recognition. The parameters of the synthesizer and the recognizer are thus optimized to maximize the success rate.

The minimization of \eq{obj1} can then be accomplished using a stochastic gradient descent algorithm, e.g.\ ADAM~\cite{Kingma15}. Each iteration of the algorithm samples a mini-batch of different bit sequences as well as different rendering layer parameter sets and updates the parameters of the synthesizer and the recognizer networks in order to minimize the loss \eq{loss} for these samples.

{\bf Practical implementation.}
As mentioned above, the components of the architecture, namely the synthesizer, the renderer, and the recognizer can be implemented as feed-forward networks. The recognizer network can be implemented as a feedforward convolutional network \cite{LeCun89} with $n$ output units. The synthesizer can use multiplicative and up-convolutional \cite{Zeiler11,Dosovitskiy15} layers, as well as element-wise non-linearities. 

\begin{figure}
    \centering
    \includegraphics[width=\textwidth]{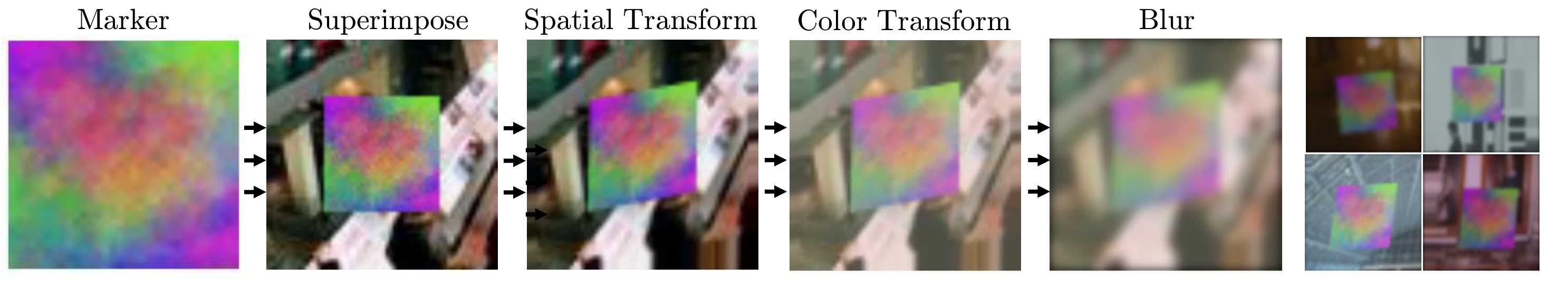}
    \label{fig:render}
    \caption{Visualizations of the rendering network $\T(M;\phi)$. For the input marker $M$ on the left the output of the network is obtained through several stages (which are all piecewise-differentiable w.r.t. inputs); on the right the outputs $\T(M;\phi)$ for several random nuisance parameters $\phi$ are shown. The use of piecewise-differentiable transforms within $T$ allows to backpropagate through $T$. }
\end{figure}

Implementing the renderer $\T(M;\phi)$ (\fig{render}) requires non-standard layers. We have implemented the renderer as a chain of layers, each introducing some ``nuisance'' transformation. We have implemented a special layer that superimposes an input over a bigger background patch drawn from a random pool of images. We use the spatial transformer layer~\cite{Jaderberg15} to implement the geometric distortion in a differentiable manner. Color shifts and intensity changes can be implemented using differentiable elementwise transformations (linear, multiplicative, gamma). Blurring associated with lens effect or motion can be simply implemented using a convolutional layer. The nuisance transformation layers can be chained resulting in a renderer layer that can model complex geometric and photometric transformations (\fig{render}).

{\bf Controlling the visual appearance.} Interestingly, we observed that under variable conditions, the optimization of \eq{obj1} results in markers that have a consistent and interesting visual texture (\fig{notexture}). Despite such style consistency, it might be desirable to control the appearance of the resulting markers more explicitly e.g.\ using some artistic prototypes. Recently, \cite{Gatys15} have achieved remarkable results in texture generation by measuring the statistics of textures using Gram matrices of convolutional maps inside deep convolutional networks trained to classify natural images. Texture synthesis can then be achieved by minimizing the deviation between such statistics of generated images and of style prototypes. Based on their approach, \cite{Ulyanov16,Johnson16} have suggested to include such deviation as a loss into the training process for deep feedforward generative neural networks. In particular, the feed-forward networks in~\cite{Ulyanov16} are trained to convert noise vectors into textures.

    We follow this line of work and augment our learning objective \eq{obj1} with the texture loss of \cite{Gatys15}. Thus, we consider a feed-forward network $\C(M;\gamma)$ that computes the result of the $t$-th convolutional layers of a network trained for large-scale natural image classification such as the \textit{VGGNet}~\cite{Simonyan14}. For an image $M$, the output $\C(M;\gamma)$ thus contains $k$ 2D channels (maps). The network $\C$ uses the parameters $\gamma$ that are pre-trained on a large-scale dataset and that are not part of our learning process. The style of an image $M$ is then defined using the following $k$-by-$k$ \textit{Gram} matrix $G(M;\gamma)$ with each element defined as: 
    \begin{equation}\label{eq:gram}
        G_{ij}(M;\gamma) = \langle\, \C_i(M;\gamma),\C_j(M;\gamma)\,\rangle\,,
    \end{equation}
    where $\C_i$ and $\C_j$ are the $i$-th and the $j$-th maps and the inner product is taken over all spatial locations. Given a prototype texture $M^0$, the learning objective can be augmented with the term: 
    \begin{equation} \label{eq:objstyle}
    f_\text{style}(\tS) = \Exp_{\b \sim U(n)}\, \left\|\,G(\S(\b;\tS);\gamma) - G(M^0;\gamma) \,\right\|^2\,.
    \end{equation}
    The incorporation of the term \eq{objstyle} forces the markers  $\S(\b;\tS)$ produced by the synthesizer to have the visual appearance similar to instances of the texture defined by the prototype $M_0$~\cite{Gatys15}.

\section{Related Work}

We now discuss the classes of deep learning methods that to the best of our understanding are most related to our approach.

Our work is partially motivated by the recent approaches that analyze and visualize pretrained deep networks by synthesizing color images evoking certain responses in these networks. Towards this end \cite{Simonyan13} generate examples that maximize probabilities of certain classes according to the network, \cite{Zeiler14} generate visual illusions that maximize such probabilities while retaining similarity to a predefined image of a potentially different class, \cite{Nguyen15} also investigate ways of generating highly-abstract and structured color images that maximize probabilities of a certain class. Finally, \cite{Mahendran15} synthesize color images that evoke a predefined vector of responses at a certain level of the network for the purpose of network inversion. Our approach is related to these approaches, since our markers can be regarded as stimuli invoking certain responses in the recognizer network. Unlike these approaches, our recognizer network is not kept fixed but is updated together with the synthesizer network that generates the marker images.

Another obvious connection are autoencoders \cite{Bengio09}, which are models trained to (1) encode inputs into a compact intermediate representation through the encoder network and (2) recover the original input by passing the compact representation through the decoder network. Our system can be regarded as a special kind of autoencoder with the certain format of the intermediate representation (a color image). Our decoder  is trained to be robust to certain class of transformations of the intermediate representations that are modeled by the rendering network. In this respect, our approach is related to variational autoencoders \cite{Kingma14} that are trained with stochastic intermediate representations and to denoising autoencoders \cite{Vincent08} that are trained to be robust to noise.

Finally, our approach for creating textured markers can be related to steganography~\cite{Petitcolas99}, which aims at hiding a signal in a \textit{carrier} image. Unlike steganography, we do not aim to conceal information, but just to minimize its ``intrusiveness'', while keeping the information machine-readable in the presence of distortions associated with printing and scanning.

\newlength{\leng}
\setlength{\leng}{1cm}

\newcommand{\set}[1]{
\addtolength{\tabcolsep}{-5pt}
\begin{tabular}{cccccc}
\includegraphics[width=\leng]{figures/#1/1.png}&
\includegraphics[width=\leng]{figures/#1/2.png}&
\includegraphics[width=\leng]{figures/#1/3.png}&
\includegraphics[width=\leng]{figures/#1/4.png}&
\includegraphics[width=\leng]{figures/#1/5.png}&
\includegraphics[width=\leng]{figures/#1/6.png}
\end{tabular}
\addtolength{\tabcolsep}{5pt}
}

\newcommand{\sset}[1]{
\includegraphics[width=\leng]{figures/#1/1.png}&
\includegraphics[width=\leng]{figures/#1/2.png}&
\includegraphics[width=\leng]{figures/#1/3.png}&
\includegraphics[width=\leng]{figures/#1/4.png}&
\includegraphics[width=\leng]{figures/#1/5.png}&
\includegraphics[width=\leng]{figures/#1/6.png}
}

\begin{figure}
\begin{tabular}{cc}
\set{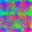}&\set{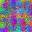}\\
64 bits, default params, C=59.9, p=99.3\%& 96 bits, low affine, C=90.2, p=99.3\%\\
\set{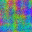}&\set{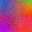}\\
64 bits, low affine $\sigma =0.05$, C=61.2, p=99.5\% & 8 bits, high blur, C=7.91, p=99.9\%\\
\set{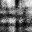}&\set{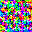}\\
32 bits, grayscale, C=27.9, p=98.3\% & 64 bits, nonlinear encoder, C=58.4, p=98.9\%\\
\set{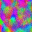}&\set{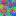}\\
64 bits, thin network, C=40.1, p=93.2\% & 64 bits, 16 pixel marker, C=56.8, p=98.5\%
\end{tabular}
\caption{Visualization of the markers learned by our approach under different circumstances shown in captions (see text for details). The captions also show the bit length, the capacity of the resulting encoding (in bits), as well as the accuracy achieved during training. In each case we show six markers: (1) -- the marker corresponding to a bit sequence consisting of $-1$, (2) -- the marker corresponding to a bit sequence consisting of $+1$, (3) and (4) -- markers for two random bit sequences that differ by a single bit, (5) and (6) -- two markers corresponding to two more random bit sequences. Under many conditions a characteristic grid pattern emerges.}
\label{fig:notexture}
\end{figure}

\section{Experiments}

Below, we present qualitative and quantitative evaluation of our approach. For longer bit sequences, the approach might not be able to train a perfect pair of a synthesizer and a recognizer, and therefore, similarly to other visual marker systems, it makes sense to use error-correcting encoding of the signal. Since the recognizer network returns the odds for each bit in the recovered signal, our approach is suitable for any probabilistic error-correction coding~\cite{Mackay03}. 

{\bf Synthesizer architectures.} For the experiments without texture loss, we use the simplest synthesizer network, which consists of a single linear layer (with a $3m^2\times{}n$ matrix and a bias vector) that is followed by an element-wise sigmoid. 
For the experiments with texture loss, we started with the synthesizer used in \cite{Ulyanov16}, but found out that it can be greatly simplified for our task. Our final architecture takes a binary code as input, transforms it with single fully connected layer and series of $3\times3$ convolutions with $2\times$ upsamplings in between.

{\bf Recognizer architectures.} Unless reported otherwise, the recognizer network was implemented as a ConvNet with three convolutional layers (96 $5\times5$ filters followed by max-pooling and ReLU), and two fully-connected layer with 192 and $n$ output units respectively (where $n$ is the length of the code). We find this architecture sufficient to successfully deal with marker encoding. In some experiments we have also considered a much smaller networks with $24$ maps in convolutional layers, and $48$ units in the penultimate layer (``thin network'').  In general, the convergence on the training stage greatly benefits from adding Batch Normalization \cite{Ioffe15} after every convolutional layer.
During our experiments with texture loss, we used VGGNet-like architecture with $3$ blocks, each consisting of two $3\times3$ convolutions and maxpooling, followed by two dense layers.

{\bf Rendering settings.} We perform a spatial transform as an affine transformation, where the 6 affine parameters are sampled from $[1,0,0,0,1,0]+\mathcal{N}(0,\sigma)$ (assuming origin at the center of the marker). The example for $\sigma = 0.1$ is shown in Fig. \ref{fig:render}. We leave more complex spatial transforms (e.g. thin plate spline \cite{Jaderberg15}) that can make markers more robust to bending for future work. Some resilience to bending can still be observed in our qualitative results.

Given an image $x$, we implement the color transformation layer as $c_1 x^{c_2} +{c_3}$, where the parameters are sampled from the uniform distribution $U[-\delta,\delta]$. As we find that printed markers tend to reduce the color contrast, we add a contrast reduction layer that transforms each value to $k x + (1-k) [0.5]$ for a random $k$. 

{\bf Quantitative measurements.} To quantify the performance of our markers under different circumstances, we report the accuracy $p$ to which our system converges during the learning under different settings (to evaluate accuracy, we threshold recognizer predictions at zero). Whenever we vary the signal length $n$, we also report the capacity of the code, which is defined as $C=n(1-H(p))$, where $H(p)=-p\log p-(1-p)\log(1-p)$ is the coding entropy.
Unless specified otherwise, we use the rendering network settings visualized in \fig{render}, which gives the impression of the variability and the difficulty of the recovery problem, as the recognizer network is applied to the outputs of this rendering network. 

\setlength{\leng}{2.2cm}
\begin{figure}
\begin{center}
\includegraphics[width=\textwidth]{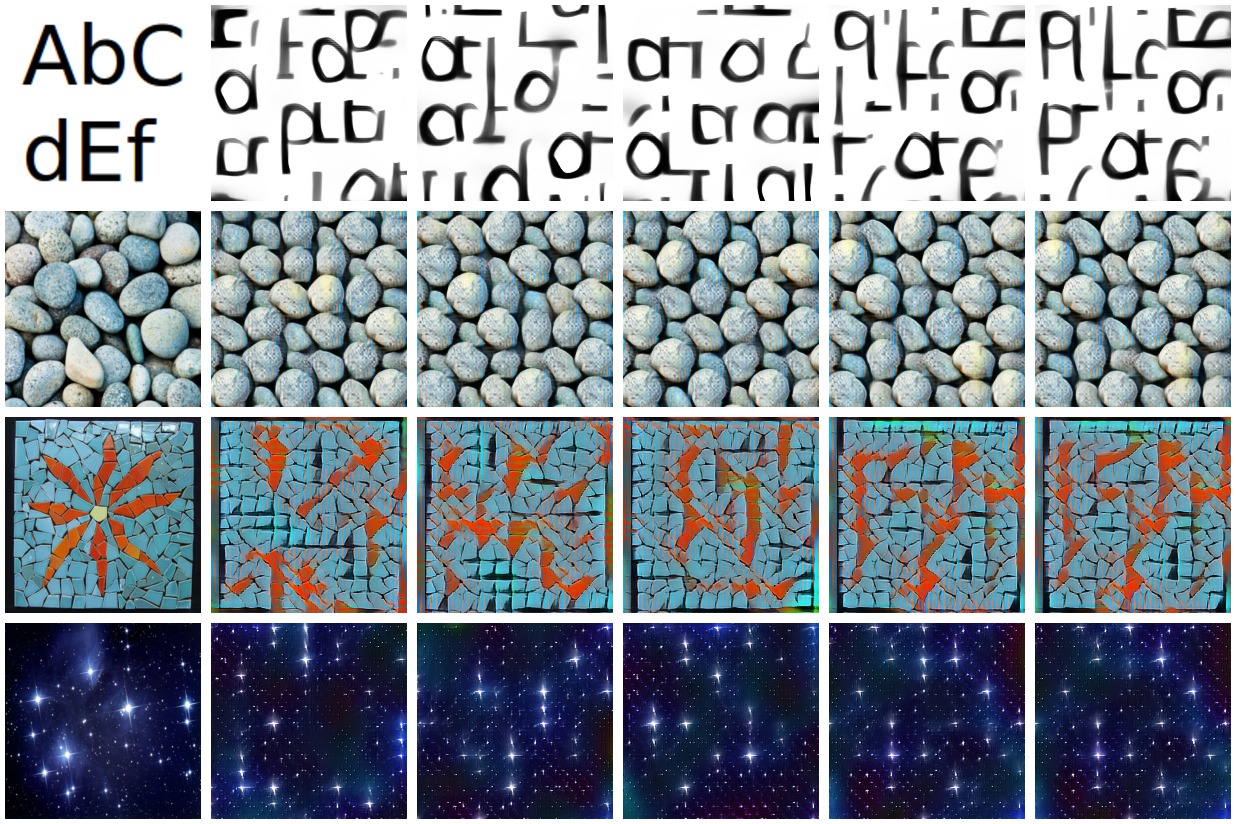}
\begin{tabularx}{\textwidth}{XXXXXX}
\centering{prototype}&
\centering{all $-1$}&
\centering{all $+1$}&
\centering{half}&
\centering{random}&
\centering{random +\\1 bit diff.}
\end{tabularx}
\addtolength{\tabcolsep}{7pt}
\end{center}
\caption{Examples of textured 64-bit marker families. The texture protototype is shown in the first column, while five remaining columns show markers for the following sequences: all $-1$, all $+1$, 32 consecutive $-1$ followed by 32 $-1$, and, finally, two random bit sequences that differ by a single bit.}
\label{fig:texture}
\end{figure}
\setlength{\leng}{1cm}

\begin{figure}
\label{fig:real}
\includegraphics[width=\textwidth]{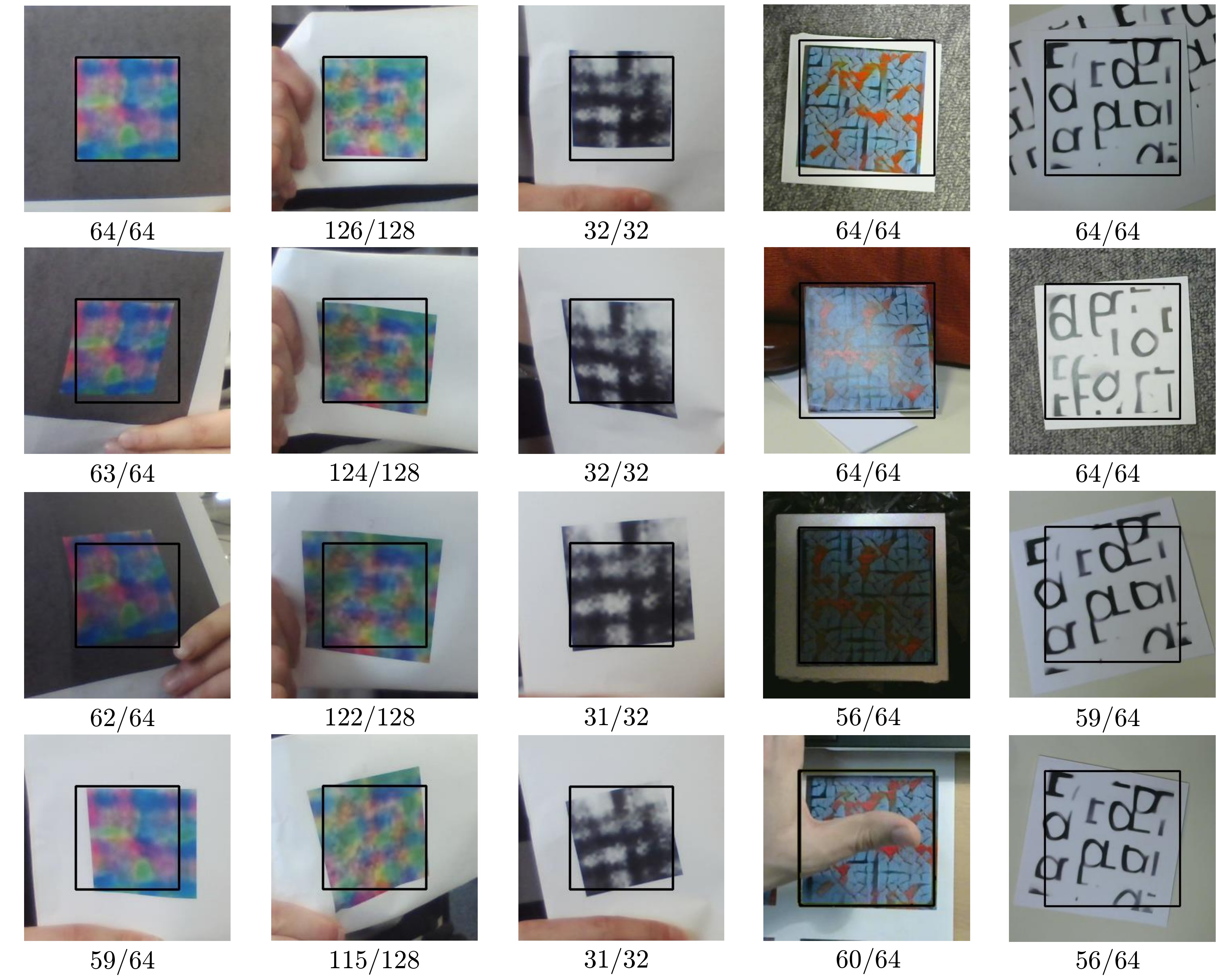}
\caption{Screenshots of marker recognition process (black box is a part of the user interface and corresponds to perfect alignment). The captions are in (number of correctly recovered bits/total sequence length) format. The rightmost two columns correspond to stylized markers. These marker families were trained with spatial variances $\sigma=0.1, 0.05, 0.1, 0.05, 0.05$ respectively. Larger $\sigma$ leads to code recovery robustness with respect to affine transformation.}
\end{figure}

\begin{figure}
\newlength\figureheight
\newlength\figurewidth
\setlength\figureheight{0.4\textwidth}
\setlength\figurewidth{0.5\textwidth}
\input{figures/bitsize.tikz}
\input{figures/markersize.tikz}
\caption{Left -- dependence of the recognition accuracy on the size of the bit string for two variants with the default networks, and one with the reduced number of maps in each convolutional layer. Reducing the capacity of the network hurts the performance a lot, while reducing spatial variation in the rendering network (to $\sigma=0.05$) increases the capacity very considerably. Right -- dependence of the recognition accuracy on the marker size (with otherwise default settings). The capacity of the coding quickly saturates as markers grow bigger. }
\label{fig:plots}
\end{figure}
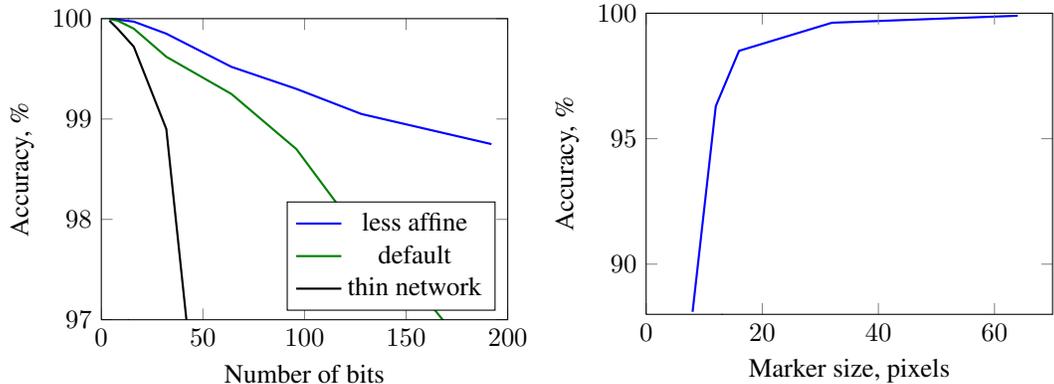

{\bf Experiments without texture loss.} The bulk of experiments without the texture loss has been performed with $m=32$ i.e.\ $32\times32$ patches (we used bilinear interpolation when printing or visualizing). The learned marker families with the base architectures as well as with its variations are shown in \fig{notexture}. It is curious to see the emergence of lattice structures (even though our synthesizer network in this case was a simple single-layer multiplicative network). Apparently, such lattices are most efficient in terms of storing information for later recovery with a ConvNet. It can also be seen how the system can adapt the markers to varying bit lengths or to varying robustness demands (e.g.\ to increasing blur or geometric distortions). We have further plotted how the quantitative performance depends on the bit length and and on the marker size in \fig{plots}.

{\bf Experiments with texture loss.} 
An interesting effect we have encountered while training synthesizer with texture loss and small output marker size is that it often ended up producing very similar patterns. We tried to tweak architecture to handle this problem but eventually found out that it goes away for larger markers.

{\bf Performance of real markers.} We also show some qualitative results that include printing (on a laser printer using various backgrounds) and capturing (with a webcam) of the markers. Characteristic results in \fig{real} demonstrate that our system can successfully recover encoded signals with small amount of mistakes. The amount of mistakes can be further reduced by applying the system with jitter and averaging the odds (not implemented here).  

Here, we aid the system by roughly aligning the marker with a pre-defined square (shown as part of the user interface). As can be seen the degradation of the results with the increasing alignment error is graceful (due to the use of affine transforms inside the rendering network at train time). In a more advanced system, such alignment can be bypassed altogether, using a pipeline that detects marker instances in a video stream and localizes their corners. Here, one can either use existing quad detection algorithms as in \cite{Olson11} or make the localization process a deep feed-forward network and include it into the joint learning in our system. In the latter case, the synthesizer would adapt to produce markers that are distinguishable from backgrounds and have easily identifiable corners. In such qualitative experiments (\fig{real}), we observe the error rates that are roughly comparable with our quantitative experiments.

{\bf Recognizer networks for QR-codes.} We have also experimented with replacing the synthesizer network with a standard QR-encoder. While we tried different settings (such as error-correction level, input bit sequence representation), the highest recognition rate we could achieve with our architecture of the recognizer network was only 85\%. Apparently, the recognizer network cannot reverse the combination of error-correction encoding and rendering transformations well. We also tried to replace both the synthesizer and the recognizer with a QR-encoder and a QR-decoder. Here we found that standard QR-decoders cannot decode QR-markers processed by our renderer network at the typical level of blur in our experiments (though special-purpose blind deblurring algorithms such as \cite{Yahyanejad10} are likely to succeed).


\section{Discussion}
In this work, we have proposed a new approach to marker design, where marker design and their recognizer are learned jointly. Additionally, an aesthetics-related term can be added into the objective. To the best of our knowledge, we are the first to approach visual marker design using optimization.

One curious side aspect of our work is the fact that the learned markers can provide an insight into the architecture of ConvNets (or whatever architecture is used in the recognizer network). In more details, they represent patterns that are most suitable for recognition with ConvNets. Unlike other approaches that e.g.\ visualize patterns for networks trained to classify natural images, our method decouples geometric and topological factors on one hand from the natural image statistics on the other, as we obtain these markers in a ``content-free'' manner\footnote{The only exception are the background images used by the rendering layer. In our experience, their statistics have negligible influence on the emerging patterns.}. 

As discussed above, one further extension to the system might be including marker localizer into the learning as another deep feedforward network.
We note that in some scenarios (e.g.\ generating augmented reality tags for real-time camera localization), one can train the recognizer to estimate the parameters of the geometric transformation in addition or even instead of the recovering the input bit string. This would allow to create visual markers particularly suitable for accurate pose estimation. 

\FloatBarrier

\small
\bibliographystyle{ieee}
\bibliography{refs}

\end{document}

%% file: figures/bitsize.tikz
%
%
%
%
\begin{tikzpicture}

\begin{axis}[
xlabel={Number of bits},
ylabel={Accuracy, \%},
xmin=0, xmax=200,
ymin=97, ymax=100,
axis on top,
legend pos=south east,
width=\figurewidth,
height=\figureheight,
legend entries={{less affine},{default}, {thin network}}
]
\addplot [thick, blue]
coordinates {
(4,99.999)
(8,99.99)
(16,99.97)
(32,99.85)
(64,99.52)
(96,99.3)
(128,99.05)
(160,98.9)
(192,98.75)

};
\addplot [thick, green!50.0!black]
coordinates {
(4,99.995)
(8,99.98)
(16,99.9)
(32,99.62)
(64,99.25)
(96,98.7)
(128,97.83)
(160,97.2)
(192,96.4)

};
\addplot [thick, black!50.0!black]
coordinates {
(4,99.98)
(8,99.90)
(16,99.72)
(32,98.9)
(64,92.75)

};

\path [draw=black, fill opacity=0] (axis cs:13,100)--(axis cs:13,100);

\path [draw=black, fill opacity=0] (axis cs:200,13)--(axis cs:200,13);

\path [draw=black, fill opacity=0] (axis cs:13,97)--(axis cs:13,97);

\path [draw=black, fill opacity=0] (axis cs:0,13)--(axis cs:0,13);

\end{axis}

\end{tikzpicture}

%% file: figures/markersize.tikz
\begin{tikzpicture}

\begin{axis}[
xlabel={Marker size, pixels},
ylabel={Accuracy, \%},
xmin=0, xmax=70,
ymin=88, ymax=100,
axis on top,
width=\figurewidth,
height=\figureheight
]
\addplot [thick, blue]
coordinates {
(8,88.1)
(12,96.3)
(16,98.5)
(32,99.62)
(64,99.9)

};
\path [draw=black, fill opacity=0] (axis cs:13,100)--(axis cs:13,100);

\path [draw=black, fill opacity=0] (axis cs:70,13)--(axis cs:70,13);

\path [draw=black, fill opacity=0] (axis cs:13,88)--(axis cs:13,88);

\path [draw=black, fill opacity=0] (axis cs:0,13)--(axis cs:0,13);

\end{axis}

\end{tikzpicture}

%% file: main.bbl
\begin{thebibliography}{10}\itemsep=-1pt

\bibitem{Belussi13}
L.~F. Belussi and N.~S. Hirata.
\newblock Fast component-based qr code detection in arbitrarily acquired
  images.
\newblock {\em Journal of mathematical imaging and vision}, 45(3):277--292,
  2013.

\bibitem{Bengio09}
Y.~Bengio.
\newblock Learning deep architectures for {AI}.
\newblock {\em Foundations and trends in Machine Learning}, 2(1):1--127, 2009.

\bibitem{Bergamasco13}
F.~Bergamasco, A.~Albarelli, and A.~Torsello.
\newblock Pi-tag: a fast image-space marker design based on projective
  invariants.
\newblock {\em Machine vision and applications}, 24(6):1295--1310, 2013.

\bibitem{Claus04}
D.~Claus and A.~W. Fitzgibbon.
\newblock Reliable fiducial detection in natural scenes.
\newblock {\em Computer Vision-ECCV 2004}, pp. 469--480. Springer, 2004.

\bibitem{Dosovitskiy15}
A.~Dosovitskiy, J.~T. Springenberg, and T.~Brox.
\newblock Learning to generate chairs with convolutional neural networks.
\newblock {\em Conf. on Computer Vision and Pattern Recognition ({CVPR})},
  2015.

\bibitem{Fiala05}
M.~Fiala.
\newblock {ARTag}, a fiducial marker system using digital techniques.
\newblock {\em Conf. Computer Vision and Pattern Recognition ({CVPR})}, v.~2,
  pp. 590--596, 2005.

\bibitem{Gatys15}
L.~Gatys, A.~S. Ecker, and M.~Bethge.
\newblock Texture synthesis using convolutional neural networks.
\newblock {\em Advances in Neural Information Processing Systems, {NIPS}}, pp.
  262--270, 2015.

\bibitem{Gatys16}
L.~A. Gatys, A.~S. Ecker, and M.~Bethge.
\newblock A neural algorithm of artistic style.
\newblock {\em Proceedings of the IEEE Conference on Computer Vision and
  Pattern Recognition,{CVPR}}, 2016.

\bibitem{Hara98}
M.~Hara, M.~Watabe, T.~Nojiri, T.~Nagaya, and Y.~Uchiyama.
\newblock Optically readable two-dimensional code and method and apparatus
  using the same, 1998.
\newblock US Patent 5,726,435.

\bibitem{Ioffe15}
S.~Ioffe and C.~Szegedy.
\newblock Batch normalization: Accelerating deep network training by reducing
  internal covariate shift.
\newblock {\em Proc. International Conference on Machine Learning, {ICML}}, pp.
  448--456, 2015.

\bibitem{Jaderberg15}
M.~Jaderberg, K.~Simonyan, A.~Zisserman, et~al.
\newblock Spatial transformer networks.
\newblock {\em Advances in Neural Information Processing Systems}, pp.
  2008--2016, 2015.

\bibitem{Johnson16}
J.~Johnson, A.~Alahi, and L.~Fei{-}Fei.
\newblock Perceptual losses for real-time style transfer and super-resolution.
\newblock {\em European Conference on Computer Vision ({ECCV})}, pp. 694--711,
  2016.

\bibitem{Kaltenbrunner07}
M.~Kaltenbrunner and R.~Bencina.
\newblock Reactivision: a computer-vision framework for table-based tangible
  interaction.
\newblock {\em Proc. of the 1st international conf. on tangible and embedded
  interaction}, pp. 69--74, 2007.

\bibitem{Kingma15}
D.~P. Kingma and J.~B. Adam.
\newblock A method for stochastic optimization.
\newblock {\em International Conference on Learning Representation}, 2015.

\bibitem{Kingma14}
D.~P. Kingma and M.~Welling.
\newblock Auto-encoding variational bayes.
\newblock {\em International Conference on Learning Representations}, 2014.

\bibitem{LeCun89}
Y.~LeCun, B.~Boser, J.~S. Denker, D.~Henderson, R.~E. Howard, W.~Hubbard, and
  L.~D. Jackel.
\newblock Backpropagation applied to handwritten zip code recognition.
\newblock {\em Neural computation}, 1(4):541--551, 1989.

\bibitem{Lo95}
C.-C. Lo and C.~A. Chang.
\newblock Neural networks for bar code positioning in automated material
  handling.
\newblock {\em Industrial Automation and Control: Emerging Technologies}, pp.
  485--491. IEEE, 1995.

\bibitem{Longacre97}
A.~Longacre~Jr and R.~Hussey.
\newblock Two dimensional data encoding structure and symbology for use with
  optical readers, 1997.
\newblock US Patent 5,591,956.

\bibitem{Mackay03}
D.~J. {MacKay}.
\newblock {\em Information theory, inference and learning algorithms}.
\newblock Cambridge university press, 2003.

\bibitem{Mahendran15}
A.~Mahendran and A.~Vedaldi.
\newblock Understanding deep image representations by inverting them.
\newblock {\em Conf. Computer Vision and Pattern Recognition ({CVPR})}, 2015.

\bibitem{Mooser06}
J.~Mooser, S.~You, and U.~Neumann.
\newblock Tricodes: A barcode-like fiducial design for augmented reality media.
\newblock {\em IEEE Multimedia and Expo}, pp. 1301--1304, 2006.

\bibitem{Nguyen15}
A.~Nguyen, J.~Yosinski, and J.~Clune.
\newblock Deep neural networks are easily fooled: High confidence predictions
  for unrecognizable images.
\newblock {\em Conf. on Computer Vision and Pattern Recognition (CVPR)}, 2015.

\bibitem{Olson11}
E.~Olson.
\newblock Apriltag: A robust and flexible visual fiducial system.
\newblock {\em Robotics and Automation (ICRA), 2011 IEEE International
  Conference on}, pp. 3400--3407. IEEE, 2011.

\bibitem{Petitcolas99}
F.~A. Petitcolas, R.~J. Anderson, and M.~G. Kuhn.
\newblock Information hiding-a survey.
\newblock {\em Proceedings of the IEEE}, 87(7):1062--1078, 1999.

\bibitem{Richardson13}
A.~Richardson and E.~Olson.
\newblock Learning convolutional filters for interest point detection.
\newblock {\em Conf. on Robotics and Automation (ICRA)}, pp. 631--637, 2013.

\bibitem{Scharstein01}
D.~Scharstein and A.~J. Briggs.
\newblock Real-time recognition of self-similar landmarks.
\newblock {\em Image and Vision Computing}, 19(11):763--772, 2001.

\bibitem{Simonyan13}
K.~Simonyan, A.~Vedaldi, and A.~Zisserman.
\newblock Deep inside convolutional networks: Visualising image classification
  models and saliency maps.
\newblock {\em arXiv preprint arXiv:1312.6034}, 2013.

\bibitem{Simonyan14}
K.~Simonyan and A.~Zisserman.
\newblock Very deep convolutional networks for large-scale image recognition.
\newblock {\em arXiv preprint arXiv:1409.1556}, 2014.

\bibitem{Ulyanov16}
D.~Ulyanov, V.~Lebedev, A.~Vedaldi, and V.~Lempitsky.
\newblock Texture networks: Feed-forward synthesis of textures and stylized
  images.
\newblock {\em Int. Conf. on Machine Learning ({ICML})}, 2016.

\bibitem{Vincent08}
P.~Vincent, H.~Larochelle, Y.~Bengio, and P.-A. Manzagol.
\newblock Extracting and composing robust features with denoising autoencoders.
\newblock {\em Int. Conf. on Machine learning ({ICML})}, 2008.

\bibitem{Woodland52}
N.~J. Woodland and S.~Bernard.
\newblock Classifying apparatus and method, 1952.
\newblock US Patent 2,612,994.

\bibitem{Yahyanejad10}
S.~Yahyanejad and J.~Str{\"o}m.
\newblock Removing motion blur from barcode images.
\newblock {\em 2010 IEEE Computer Society Conference on Computer Vision and
  Pattern Recognition-Workshops}, pp. 41--46. IEEE, 2010.

\bibitem{Zeiler14}
M.~D. Zeiler and R.~Fergus.
\newblock Visualizing and understanding convolutional networks.
\newblock {\em Computer vision--ECCV 2014}, pp. 818--833. Springer, 2014.

\bibitem{Zeiler11}
M.~D. Zeiler, G.~W. Taylor, and R.~Fergus.
\newblock Adaptive deconvolutional networks for mid and high level feature
  learning.
\newblock {\em Int. Conf. on Computer Vision (ICCV)}, pp. 2018--2025, 2011.

\end{thebibliography}
